\documentclass{article}

\usepackage{amsmath,amsfonts,bm}

\def\1{\bm{1}}

\def\vu{{\bm{u}}}

\def\mE{{\bm{E}}}

\def\mL{{\bm{L}}}

\def\mX{{\bm{X}}}

\DeclareMathAlphabet{\mathsfit}{\encodingdefault}{\sfdefault}{m}{sl}
\SetMathAlphabet{\mathsfit}{bold}{\encodingdefault}{\sfdefault}{bx}{n}

\DeclareMathOperator*{\argmax}{arg\,max}
\DeclareMathOperator*{\argmin}{arg\,min}

\usepackage{microtype}
\usepackage{graphicx}
\usepackage{amsmath}
\usepackage{amssymb}
\usepackage{amsthm}
\usepackage{multirow}
\usepackage{makecell}
\newtheorem{theorem}{Theorem}[section]

\usepackage{subfigure}
\usepackage{booktabs} %
\usepackage[dvipsnames]{xcolor}
\usepackage{mathtools}
\usepackage{bbm}

\usepackage{hyperref}

\usepackage[accepted]{icml2020}

\usepackage{xcolor}
\definecolor{Orange}{RGB}{244, 101, 66}
\definecolor{Red}{RGB}{255, 3, 13}
\definecolor{Plum}{RGB}{142, 69, 133}

\icmltitlerunning{An EM Approach to Non-autoregressive Conditional Sequence Generation}

\begin{document}

\twocolumn[
\icmltitle{An EM Approach to Non-autoregressive Conditional Sequence Generation}

\icmlsetsymbol{equal}{*}

\begin{icmlauthorlist}
\icmlauthor{Zhiqing Sun}{cmu}
\icmlauthor{Yiming Yang}{cmu}
\end{icmlauthorlist}

\icmlaffiliation{cmu}{Carnegie Mellon University, Pittsburgh, PA 15213 USA}

\icmlcorrespondingauthor{Zhiqing Sun}{zhiqings@cs.cmu.edu}

\icmlkeywords{Machine Translation, EM Algorithm}

\vskip 0.3in
]

\printAffiliationsAndNotice{}  %

\begin{abstract}

Autoregressive (AR) models have been the dominating approach to conditional sequence generation, but are suffering from the issue of high inference latency.  Non-autoregressive (NAR) models have been recently proposed to reduce the latency by generating all output tokens in parallel but could only achieve inferior accuracy compared to their autoregressive counterparts, primarily due to a difficulty in dealing with the multi-modality in sequence generation.  This paper proposes a new approach that jointly optimizes both AR and NAR models in a unified Expectation-Maximization (EM) framework. In the E-step, an AR model learns to approximate the regularized posterior of the NAR model. In the M-step, the NAR model is updated on the new posterior and selects the training examples for the next AR model. This iterative process can effectively guide the system to remove the multi-modality in the output sequences. To our knowledge, this is the first EM approach to NAR sequence generation. We evaluate our method on the task of machine translation. Experimental results on benchmark data sets show that the proposed approach achieves competitive, if not better, performance with existing NAR models and significantly reduces the inference latency.

\end{abstract}

\section{Introduction} \label{sec:introduction}

State-of-the-art conditional sequence generation models~\citep{bahdanau2014neural,gehring2017convolutional,vaswani2017attention} 
typically rely on an AutoRegressive (AR) factorization scheme to produce the output sequences. Denoting by $x=(x_1,\dots,x_T)$ an input sequence of length $T$, and by $y=(y_1,\dots,y_{T'})$ a target sequence of length $T'$, the conditional probability of $y$ given $x$ is factorized as:
\begin{equation} \label{eq:1}
    p^{AR}(y|x) = \prod_{i=1}^{T'}p(y_i | x, y_1, y_2, \dots, y_{i-1}).
\end{equation}
As such a sequential factorization cannot take the full advantage of parallel computing, it yields high \emph{inference latency} as a limitation.

Recently, Non-AutoRegressive (NAR) sequence models~\citep{gu2017non,lee2018deterministic} are proposed to tackle the problem of 
inference latency, by removing the sequential dependencies among the output tokens as:
\begin{equation} \label{eq:2}
    p^{NAR}(y|x) = p(T'|x)\prod_{i=1}^{T'}p(y_i | x, T').
\end{equation}
This formulation allows each token to be decoded in parallel and hence brings a significant reduction of the inference latency.  However, NAR models also suffer from the conditional independence assumption among the output tokens, and usually do not perform as well as their AR counterparts. Such a performance gap is particularly evident when the output distributions exhibit a multi-modality phenomenon~\cite{gu2017non}.
which means that the input sequence can be mapped to multiple correct output sequences. Such a multi-modal output distribution cannot be represented as the product of conditionally independent distributions for each position in NAR models (See \ref{sec:multimodality} for a detailed discussion).

How to overcome the multi-modality issue has been 
a central focus in recent efforts for improving NAR models.
A standard approach is to use sequence-level knowledge distillation~\citep{hinton2015distilling,kim2016sequence}, which means to replace the target part $y$ of each training instance $(x, y)$ with the system-predicted $\hat{y}$ from a pre-trained AR model (a.k.a. the "teacher model").
Such a replacement strategy removes the one-to-many mappings from the original dataset.  The justification for doing so is that in practice we do not really need the sequence generation models to mimic a
diverse output distribution for sequence generation tasks such as machine translation\footnote{For example, Google Translate only provide one translation for the input text.} and text summarization. Such a knowledge distillation strategy has shown to be effective for improving the performance of NAR models in multiple studies~\citep{gu2017non,kaiser2018fast,li2019hint,ma2019flowseq,sun2019fast,ghazvininejad2019mask,gu2019levenshtein}.

\begin{figure}
	\centering
	\includegraphics[width=0.9\linewidth]{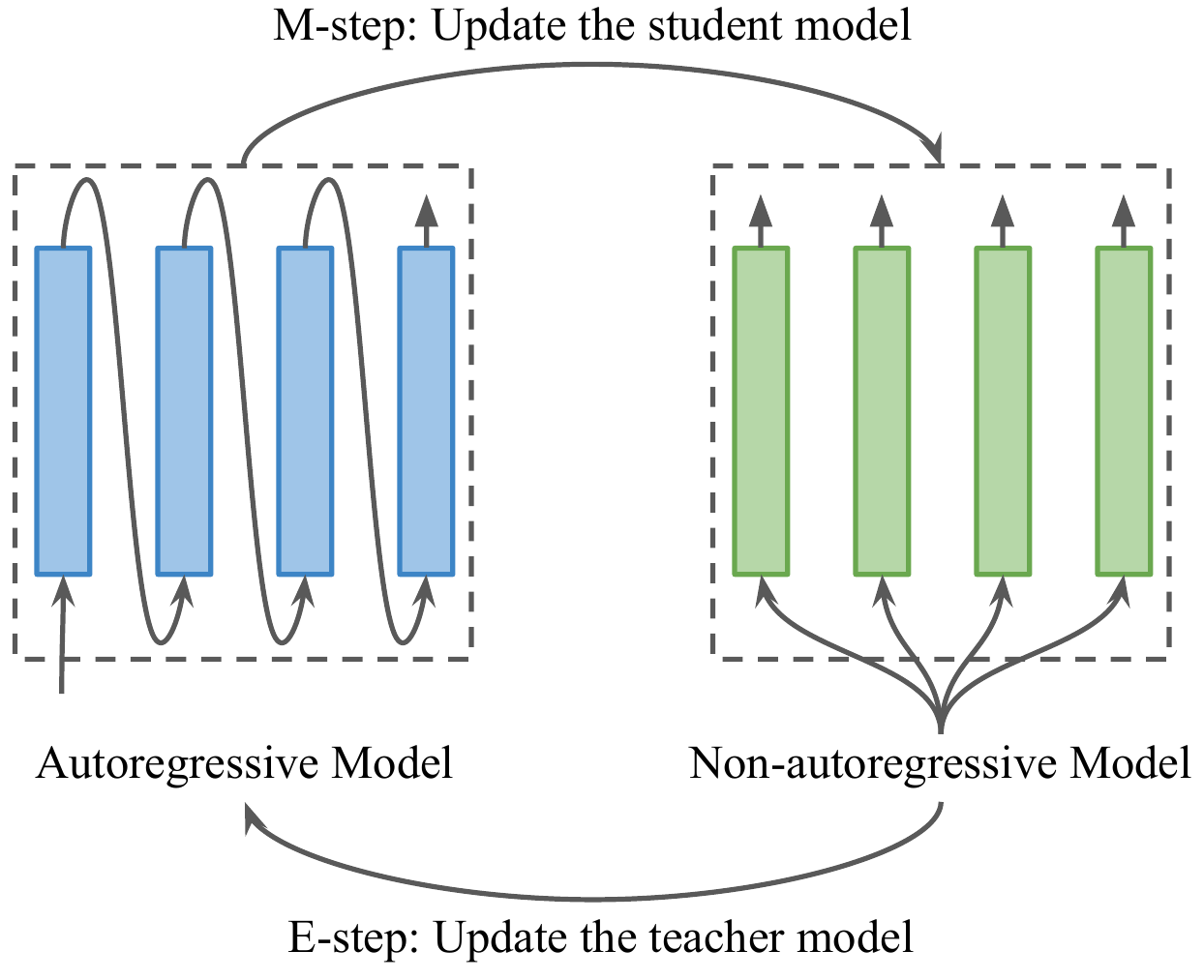}
	\caption{The proposed framework:  The AR and NAR models are jointly optimized by alternating between a E-step and a M-step. See Sec.~\ref{sec:em} for the detailed explanation.}
	\label{fig:framework}
\end{figure}

We want to point out that in all the NAR methods with knowledge distillation, the teacher AR model is pre-trained only once on ground-truth data  and then is used to generate the output training examples for the NAR model.
We argue that such a single-pass knowledge distillation process may not be sufficient for optimizing the NAR model as sequence $\hat{y}$ predicted by the AR model cannot be perfect. More importantly, it is
not necessarily the best choice for alleviating the multi-modality problem in the NAR model.  In other words, without knowing how the choice of $\hat{y}$ by the AR model would effect the training process in the NAR model, the current knowledge distillation approach is unavoidably sub-optimal.

To address this fundamental limitation, we propose a novel Expectation-Maximization (EM) approach to
the training of NAR models, where both the teacher (an AR model) and the student (a NAR model) are helping each other in a closed loop, and the iterative updates of the models are guaranteed to converge to a local optimum. This approach gives an extra power to knowledge distillation between AR and NAR models, and is the first EM approach to NAR models, to our knowledge.  Fig.~\ref{fig:framework} illustrates our new framework. 
In addition, we develop a principled plug-and-play decoding module for effective removal of word duplication in the model's output.
Our experiments on three machine translation benchmark datasets show that the proposed approach achieves competitive performance as that of the best NAR models in terms of prediction accuracy, %
and reduces the inference latency significantly.

\section{Related Work} \label{sec:related}
Related work can be divided into two groups, i.e., the non-autoregressive methods for conditional sequence generation, and the various approaches to knowledge distillation in non-autoregressive models.

Recent work on non-autoregressive sequence generation has developed ways to address the multi-modality problem.
Several work try to design better training objectives~\citep{shao2019minimizing,wei2019imitation} or regularization terms~\citep{li2019hint,wang2019non,guo2018non}.
Other methods focus on direct modeling of
multi-modal target distributions via hidden variables~\citep{gu2017non,kaiser2018fast,ran2019guiding,ma2019flowseq} or sophisticated output structures~\citep{libovicky2018end,sun2019fast}.
There are also a few recent work~\citep{lee2018deterministic,stern2019insertion,ghazvininejad2019mask,gu2019levenshtein} focusing on a multiple-pass iterative-refinement process to generate the final outputs, where the first pass produce an initial output sequence and the following passes refine the sequence iteratively in the inference phase.

As for knowledge distillation in NAR models, 
\citet{gu2017non} is the first effort to use knowledge distillation~\citep{hinton2015distilling,kim2016sequence}.
Recently, \citet{zhou2019understanding} 
analyzed why knowledge distillation would reduce the complexity of datasets and hence be helpful in the training of NAR models. 
They also used Born-Again networks (BANs)~\citep{furlanello2018born} to produce simplified training data for NAR models. 

All the above methods take a pre-trained AR model as the teacher model for knowledge distillation; none of them iteratively updates the teacher model based on the feedback from  (or the measured performance of) the NAR model.  This is the fundamental difference between existing work and our proposed EM approach in this paper.

\section{Problem Definition} \label{sec:problem}

\subsection{Conditional Sequence Generation}

Let us describe the problem of conditional sequence generation in the context of machine translation and use the terms of ``sequence'' and ``sentence'', ``source'' and ``input'', ``target'' and ``output''  interchangeably. %
We use $x$ and $y$ to denote the source and target sentences, $x_i$ to indicate the $i$-th token in $x$, and $\mathcal{X}=\{x^1, x^2,\dots,x^N\}$ and  $\mathcal{Y}=\{y^1, y^2,\dots,y^N\}$ to be a parallel dataset of $N$ sentence pairs in the source and target languages, respectively.

The training of both AutoRegressive (AR) and  Non-AutoRegressive (NAR) sequence generation models is performed via likelihood maximization over parallel data $(\mathcal{X},\mathcal{Y})$ as:
\begin{align}
    \phi^*&= \argmax_{\phi} \mathbb{E}_{(x, y) \sim (\mathcal{X},\mathcal{Y})}\log p^{AR}(y|x; \phi), \\
    \theta^*&= \argmax_{\theta} \mathbb{E}_{(x, y) \sim (\mathcal{X},\mathcal{Y})} \log p^{NAR}(y|x; \theta),
\end{align}
where $p^{AR}$ and $p^{NAR}$ are defined in Eq.~\ref{eq:1} and Eq.~\ref{eq:2}; $\phi$ and $\theta$ are the parameters of the AR and NAR models, respectively.

\subsection{Instance-level \& Corpus-level Multi-modality} \label{sec:multimodality}

There are two different but not mutually exclusive definitions for the concept of multi-modality (a.k.a., translation uncertainty in machine translation terminology).

\citet{gu2017non} define the multi-modality problem as the existence of one-to-many mappings in the parallel data, which we refer to as the \emph{instance-level} multi-modality.  Formally, given the parallel data $(\mathcal{X}, \mathcal{Y})$, if there exist sentences $i$ and $j$ such that $x^i = x^j$ but $y^i \not= y^j$, we say that $(\mathcal{X}, \mathcal{Y})$ contains instance-level multi-modality.

In contrast, \citet{zhou2019understanding} use the conditional entropy to quantify the translation uncertainty, which we refer as the \emph{corpus-level} multi-modality. It requires the 
use of an additional alignment model in calculating this quantity.
In this paper, we want to avoid the requirement for
external alignment tools.  Thus, we directly use  training-set likelihood of the NAR model as a measure of corpus-level multi-modality.
Formally, the Corpus-level Multi-modality (CM) of parallel data $(\mathcal{X}, \mathcal{Y})$ is defined as:
\begin{align}
    &\mathrm{CM}_{\mathcal{X}}(\mathcal{Y}) = \mathbb{E}_{(x, y) \sim (\mathcal{X}, \mathcal{Y})} \left[- \log p^{NAR}(y|x; \theta^*)\right], \label{eq:3}\\
    &\theta^* = \argmax_{\theta} \mathbb{E}_{(x, y) \sim (\mathcal{X},\mathcal{Y})} \log p^{NAR}(y|x; \theta).
\end{align}
To make this metric comparable across different data sets, we further define the Normalized Corpus-level Multi-modality (NCM) as:
\begin{equation} \label{eq:4}
    \mathrm{NCM}_{\mathcal{X}}(\mathcal{Y}) = \frac{\mathbb{E}_{(x, y) \sim (\mathcal{X}, \mathcal{Y})} \left[- \log p^{NAR}_{\mathcal{M}}(y|x; \theta^*)\right]}{\mathbb{E}_{y \sim \mathcal{Y}} \left[ |y| \right]}
\end{equation}
where $|y|$ denotes the length of output sequence $y$.

\newtheorem{prop}{Proposition}[section]

\section{Properties of NAR Models} \label{sec:how}

How powerful are NAR models? Are they as expressive as AR models?
Our answer is both yes and no. On one hand, it is easy to see that NAR models can only capture distributions that can be factorized into conditional independent parts. On the other hand, we will show in the next that if the instance-level multi-modality can be removed from the  training data (e.g., via sequence-level knowledge distillation), then NAR models can be as powerful just as AR models.

\subsection{Theoretical expressiveness of NAR models}

Let us focus on the expressive power of NAR models when the instance-level multi-modality is removed, that is, when there are only one-to-one and many-to-one mappings in training examples. More specifically, we consider the ability of the vanilla Non-Autoregressive Transformer (NAT) models \citep{gu2017non,vaswani2017attention} in approximating arbitrary continuous $\mathbb{R}^{d\times n} \rightarrow \mathbb{R}^{d\times m}$ sequence-to-sequence single-valued functions, where $n$ and $m$ are the input and output sequence length, and $d$ is the model dimension.

Given the definition of a distance between two functions $f_1, f_2: \mathbb{R}^{d\times n} \rightarrow \mathbb{R}^{d\times m}$ as:
\begin{equation}
    d_p(f_1, f_2) = \left(\int \|f_1(\mathbf{X}) - f_2(\mathbf{X}) \|^p_p d \mathbf{X} \right)^{1/p}.
\end{equation}
where $p \in [1, \infty)$. We can make the following statement:
\begin{theorem} \label{the1}
Let $1 \le p < \infty$ and $\epsilon > 0$, then for any given continuous sequence-to-sequence function $f: \mathbb{R}^{d\times n} \rightarrow \mathbb{R}^{d\times m}$, there exists a non-autoregressive Transformer network $g$ such that $d_p(f, g) \le \epsilon$.
\end{theorem}

This theorem is a corollary of the Theorem 2 in \citet{yun2020are}. For completeness, we provide the formal theorem with proof in the appendix.

\subsection{What limits the success of NAR models in practice?}

Theorem \ref{the1} shows that for any sequence-to-sequence dataset containing no instance-level multi-modality, we can always find a good NAT model to fit this dataset. However, in reality, it is still a big challenge for NAT models to fit the distilled deterministic training data very well.

The gap between theory and practice is due to the fact that in theory we may use as many Transformer layers as needed, but in reality, there are only a few layers (e.g., 6 layers) in the Transformer model. This would greatly restrict the model capacity of real NAR models.

To further understand the limitation let us examine the following two hypotheses:
\begin{itemize}
    \item  The NAT model intrinsically cannot accurately produce very long output sequences when it has only a few Transformer layers.
    \item %
    The corpus-level multi-modality in data makes it hard for NAT models to deal with (i.e., to memorize the ``mode'' in the output for each input).
\end{itemize}

\setlength{\tabcolsep}{4pt}
\begin{table}[t]
\centering
\small
\caption{Toy examples illustrating the two types of synthetic experiments.} %
\vspace{5px}
\begin{tabular}{c c c c} 
\hline
 & source & & target \\
\hline
\multirow{3}{*}{\makecell{Experiment I}}& 
$\mathtt{2\ 1\ 4\ 3}$ & $\rightarrow$ & $\mathtt{2\ 2\ 1\ 4\ 4\ 4\ 4\ 3\ 3\ 3}$

\\

&$\mathtt{2\ 2\ 3}$  & $\rightarrow$ &  $\mathtt{2\ 2\ 2\ 2\ 3\ 3\ 3}$\\

& $\mathtt{2\ 1\ 5}$ & $\rightarrow$ &  $\mathtt{2\ 2\ 1\ 5\ 5\ 5\ 5}$\\

\hline

\multirow{3}{*}{\makecell{Experiment II}}&
 $\mathtt{2\ 1\ 4\ 3}$  & $\rightarrow$ &  $\mathtt{0\ 2\ 2\ 1\ 4\ 4\ 4\ 4\ 3\ 3\ 3\ 0\ 0\ 0}$
\\

& $\mathtt{2\ 2\ 3}$ & $\rightarrow$ &  $\mathtt{0\ 0\ 0\ 2\ 2\ 2\ 2\ 3\ 3\ 3\ 0}$\\

& $\mathtt{2\ 1\ 5}$ & $\rightarrow$ &  $\mathtt{2\ 2\ 1\ 5\ 5\ 5\ 5\ 5\ 0\ 0\ 0\ 0}$\\

\hline
\end{tabular}
\label{tab:synthetic_example}
\end{table}
\setlength{\tabcolsep}{6pt}

\setlength{\tabcolsep}{4pt}
\begin{table}[t]
\centering
\small
\caption{The accuracy in whole-sentence matching of the AR and NAR models over 1000 synthetic examples.} %
\vspace{5px}
\begin{tabular}{c c c c c} 
\hline
 & \multicolumn{2}{c}{Experiment I} & \multicolumn{2}{c}{Experiment II} \\
Models & AR & NAR & AR & NAR \\
\hline
Accuracy(\%) & 99.9 & 95.7 & 99.8 & 00.0\\
\hline
\end{tabular}
\label{tab:synthetic}
\end{table}
\setlength{\tabcolsep}{6pt}

These hypotheses focus on two different reasons that might cause the poor performance of NAR models. In order to verifying which one is true, we design two types of synthetic data and experiments.

In Experiment I, a synthetic translation dataset is constructed as follows: The source and target sides share the same vocabulary of $\{\mathtt{1},\mathtt{2},\mathtt{3},\mathtt{4},\mathtt{5}\}$. $\mathtt{1}$, $\mathtt{2}$, $\mathtt{3}$, and $\mathtt{4}$, and $\mathtt{5}$ are translated into $\mathtt{1}$, $\mathtt{2}\ \mathtt{2}$, $\mathtt{3}\ \mathtt{3}\ \mathtt{3}$, $\mathtt{4}\ \mathtt{4}\ \mathtt{4}\ \mathtt{4}$, and $\mathtt{5}\ \mathtt{5}\ \mathtt{5}\ \mathtt{5}\ \mathtt{5}$, respectively. 
The translation is deterministic, i.e., no multi-modality in the resulted parallel dataset.
In Experiment II, we randomly insert four $0$ in the front or the back of each target sentence in the 1st dataset. In other words, the source-to-target translation in the 2nd dataset is non-deterministic and hence is with corpus-level multi-modality. In addition, we filter the source data in Experiment II to make sure that there is no instance-level multi-modality in this dataset. The toy examples illustrating the two types of datasets can be found in Tab.~\ref{tab:synthetic_example}.
Following such rules we randomly generated 2,000,000 sentences for training and 1000 sentences for testing; both the training and testing \textit{source} sentences have the length of 30.

We trained both the AR transformer and the NAR Transformer on these synthetic datasets, each of
which consists of a 3-layer encoder and a 3-layer decoder. The detailed model settings can be found in the appendix. In our  evaluation we used the ground-truth lengths for the decoding of both the AR and NAR Transformers; the whole-sentence matching accuracy of those models are listed in Tab.~\ref{tab:synthetic}.

The results in Experiment I show that both the autoregressive Transformer and the non-autoregressive Transformer can achieve an high accuracy of $99.9\%$  and $95.7\%$, respective,
when the training data do not have the multi-modality. In contrast, the results of Experiment II show that the non-autoregressive Transformer model failed completely on the synthetic dataset with corpus-level multi-modality.  The sharp contrast in these synthetic experiments indicates that  the real problem with NAR models is indeed due to the corpus-level multi-modality issue.

\section{Proposed Method} \label{sec:method}

Let us formally introduce our EM approach to addressing the multi-modality issue in NAR models, followed by a principled decoding module for effective removal of word duplication in the predicted output.

\subsection{The EM Framework} \label{sec:em}

With the definition of the corpus-level multi-modality (i.e., $\mathrm{CM}$ in Eq.~\ref{eq:3}), we consider how to reduce this quantity for the better training of NAR models. Formally,
given source data $\mathcal{X}$, we want to find target data $\mathcal{Y}^*$ that satisfies the following property:
\begin{align*}
\mathcal{Y}^*
&= \argmin_{\mathcal{Y}} \mathrm{CM}_{\mathcal{X}}(\mathcal{Y})\\
&= \argmin_{\mathcal{Y}} \mathbb{E}_{(x, y) \sim (\mathcal{X}, \mathcal{Y})} \left[-\log p^{NAR}(y|x; \theta^*)\right].
\end{align*}
However, there can be many trivial solutions for $\mathcal{Y}^*$. For example, we can simply construct a dataset with no variation in the output to achieve zero corpus-level multi-modality. To avoid triviality, we may 
further apply a constraint to $\mathcal{Y}^*$. This leads us to the posterior regularization framework.

Posterior Regularization~\citep{ganchev2010posterior} is a probabilistic framework for structured, weakly supervised learning.
In this framework, we can re-write our objective as following:
\begin{equation*}
    \mathcal{L}_{\mathcal{Q}}(\theta) =  \min_{q\in \mathcal{Q}} \mathbb{KL}(q(\mathcal{Y})\|p^{NAR}(\mathcal{Y}|\mathcal{X}; \theta)),
\end{equation*}
where $q$ is the posterior distribution of $\mathcal{Y}$ and $\mathcal{Q}$ is a constraint posterior set that controls the quality of the parallel data given by:
\begin{equation*} \label{eq:pr}
    \mathcal{Q}= \{q(\mathcal{Y}): \mathbb{E}_{\mathcal{Y} \sim q} \left[\mathrm{Q}_{\mathcal{X}}(\mathcal{Y})\right] \ge \mathbf{b}\},
\end{equation*}
where $\mathrm{Q}$ is a metric for quality mapping from $(\mathcal{X},\mathcal{Y})$ to $\mathbb{R}^N$ in the training set and $\mathbf{b}$ is a bound vector.

For sequence generation tasks, there are many corpus-level quality metrics, such as BLEU~\citep{papineni2002bleu} and ROUGE~\citep{lin2003automatic}. However, they are known to be inaccurate for measuring the quality of single sentence pairs. Thus, 
we use the likelihood score of a pre-trained AR model as a more reliable quality metric:
\begin{equation*}
    \left[\mathrm{Q}_{\mathcal{X}}(\mathcal{Y})\right]_{i} = \mathrm{Q}_{x^i}(y^i) = \log p^{AR}(y^i|x^i; \phi^1),
\end{equation*}
where $\phi^1$ denotes that the %
AR model trained on the original ground-truth dataset.

Given the posterior regularization likelihood $\mathcal{L}_{Q}(\theta)$, 
we use the EM algorithm \citep{mclachlan2007algorithm,ganchev2010posterior}
to optimize it. In the E-step (a.k.a,the inference procedure), the goal is to fix $p^{NAR}$ and update the posterior distribution:
\begin{equation*}
    q^{t+1} = \argmin_{q\in Q} \mathbb{KL}(q(\mathcal{Y})\|p^{NAR}(\mathbf{\mathcal{Y}}|\mathcal{X}; \theta^t)), \label{em1}
\end{equation*}
In the M-step (a.k.a., learning procedure), we will fix $q(\mathcal{Y})$ and update $\theta$ to maximize the expected log-likelihood:
\begin{equation*}
    \theta^{t+1} = \argmax_{\theta}  \mathbb{E}_{q^{t+1}}[\log p^{NAR}(\mathcal{Y}|\mathcal{X}; \theta)], \label{em2}
\end{equation*}
Next, we introduce the details of the E-step and the M-step in our framework.

\paragraph{Inference Procedure}

The E-step aims to compute the posterior distribution $q(\mathcal{Y})$ that minimizes the KL divergence between $q(\mathcal{Y})$ and $p^{NAR}(\mathcal{Y}|\mathcal{X})$.
\citet{ganchev2010posterior} show that for graphical models, $q(\mathcal{Y})$ can be efficiently solved in its dual form. Specifically, the primal solution $q^*$ is given in terms of the dual solution $\boldsymbol{\lambda}^*$ by:
\begin{align*}
    q^*(\mathcal{Y})
    &\propto p^{NAR}(\mathcal{Y}|\mathcal{X}; \theta^t) \exp\{\boldsymbol{\lambda}^* \cdot \mathrm{Q}_{\mathcal{X}}(\mathcal{Y})\}\\
    &\propto  \prod_{i=1}^{N} \left[p^{NAR}(y^i|x^i; \theta^t) \left(p^{AR}(y^i|x^i; \phi^0)\right)^{\lambda^{*}_i} \right],
\end{align*}
However, a problem here, as pointed out by \citet{zhang2018prior}, is that it is hard to specify the hyper-parameter $\mathbf{b}$ to effectively bound the
expectation of the features 
for neural models. Besides, even when $\mathbf{b}$ is given, calculating $\boldsymbol{\lambda}^*$ is still intractable for neural models.
Therefore, in this paper, we 
introduce another way to compute $q(\mathcal{Y})$.

We first factorize $q(\mathcal{Y})$ as the product of $\{q(y^i)\}$, and then follow the idea of amortized inference~\citep{gershman2014amortized} to parameterize $q(y^i)$ with an AR sequence generation model:
\begin{equation*}
    q(\mathcal{Y}) = \prod_{i=1}^N p^{AR}(y^i|x^i; \phi). \label{eq:mf2}
\end{equation*}
The E-step can thus be re-written as follows:
\begin{align*}
    \phi^{t+1} 
    &= \argmin_{\phi \in \mathcal{Q}'} \mathbb{E}_{x \sim \mathcal{X}}\mathbb{KL}(p^{AR}(y|x; \phi)\|p^{NAR}(y|x; \theta^t)).
\end{align*}
where the new constraint posterior set $\mathcal{Q}'$ is defined as
\begin{equation*}
    \left\{\phi: \mathbb{E}_{p^{AR}(\mathcal{Y}|\mathcal{X}; \phi)}\left[\mathrm{CQ}_{\mathcal{X}}(\mathcal{Y})\right] \ge b \right\}.
\end{equation*}
We %
further apply the REINFORCE algorithm~\citep{williams1992simple} to estimate the gradient of $\mathcal{L}_{\mathcal{Q}}$ w.r.t. $\phi \in \mathcal{Q}'$:
\begin{equation*} \label{eq:inference}
\begin{split}
    & \nabla_{\phi} \mathcal{L}_{\mathcal{Q}} = 
    \mathbb{E}_{x \sim \mathcal{X}} \mathbb{E}_{y \sim p^{AR}(y|x; \phi)} \\
    & \left(- \log \frac{p^{NAR}(y|x;\theta^t)}{p^{AR}(y|x;\phi)}  \nabla_{\phi} \log p^{AR}(y|x;\phi) \right).
\end{split}
\end{equation*}
This can be %
intuitively viewed as to construct
 a weighted pseudo training dataset $(\mathcal{X}^{t+1},\mathcal{Y}^{t+1})$, where the training examples are sampled at random from $p^{AR}(y|x;\phi)$ and weighted by $\log \frac{p^{NAR}(y|x;\theta^t)}{p^{AR}(y|x;\phi)}$.

In practice, we find that there are two problems in implementing this algorithm: One is that sampling from $p^{AR}(y|x;\phi)$ is very inefficient; the other is that the constraint $\phi \in \mathcal{Q}'$ cannot be guaranteed. Therefore, we instead use a heuristic way when constructing the pseudo training dataset $(\mathcal{X}^{t+1},\mathcal{Y}^{t+1})$: We first follow \citet{wu2018study} to replace the inefficient sampling process with beam search \citep{sutskever2014sequence} on $p^{AR}(y|x; \phi^t)$, and then filter out the candidates that doesn't satisfy the following condition:
\begin{equation*}
    \mathrm{Q}_{x}(y) \ge \hat{b}_i,
\end{equation*}
where $\hat{b}_i$ is a newly introduced pseudo bound that can be empirically set by early stopping.
In this way, we control the quality of $p^{AR}(y|x;\phi^{t+1})$ by manipulating the quality of its training data. Finally, we choose the ones with the highest $p^{AR}(y|x; \phi^t) \log \frac{p^{NAR}(y|x;\theta^{t})}{p^{AR}(y|x; \phi^t)}$ score as the training examples in $\mathcal{Y}^{t+1}$, and $\mathcal{X}^{t+1}$ is merely a copy of $\mathcal{X}$.

In each E-step, in principle, we should let $\phi^{t+1}$ converge under the current NAR model.
Although this can be achieved by constructing the pseudo datasets and train the AR model for multiple times, it is practically prohibitive due to the expensive training cost. We, therefore, use only a single update iteration of the AR model in the inner loop of each E-step.

\paragraph{Learning Procedure}
In the M-step, we seek to learn the parameters $\theta^{t+1}$ with the parameterized posterior distribution $p^{AR}(\mathcal{Y}|\mathcal{X}; \phi^{t+1})$.
However, directly sampling training examples from the AR model will cause the instance-level multi-modality problem. Therefore, we apply sequence-level knowledge distillation~\citep{kim2016sequence} to solve this problem, that is, we only use the targets with maximum likelihood in the AR model to train the NAR model:
\begin{equation*} \label{eq:learning}
\theta^{t+1} = \argmax_{\theta}  \mathbb{E}_{(x, y) \sim (\mathcal{X}, \hat{\mathcal{Y}}^{t+1})} \log p^{NAR}(y|x; \theta).
\end{equation*}
where $\hat{\mathcal{Y}}^{t+1}$ denotes the training examples produced by the AR model $p^{AR}(\mathcal{Y}|\mathcal{X}; \phi^{t+1})$.

\paragraph{Joint Optimization}
We first pre-train an AR teacher model on the ground-truth parallel data as $p^{AR}(y|x; \phi^{1})$.
Then we alternatively optimize $p^{NAR}$ and $p^{AR}$ until convergence.
We summarize the optimization algorithm in Alg. \ref{alg:example}.
In our EM method, the AR and NAR models are jointly optimized to reduce the corpus-level multi-modality.

\begin{algorithm}[t]
   \caption{An EM approach to NAR models}
   \label{alg:example}
\begin{algorithmic}
   \STATE {\bfseries Input:} Parallel training dataset $(\mathcal{X}, \mathcal{Y})$
   \STATE $t = 0$
   \STATE Pre-train $p^{AR}(y|x; \phi^1)$ on $(\mathcal{X}, \mathcal{Y})$.
   \WHILE{not converged}
        \STATE $t = t + 1$
        \STATE $\boxdot$ \textbf{M-Step: Learning Procedure}
        \STATE Construct the distillation dataset $\hat{\mathcal{Y}}^{t}$ with $p^{AR}(y|x; \phi^{t})$.
        \STATE Train $p^{NAR}(y|x;\theta^{t})$ on $(\mathcal{X}, \hat{\mathcal{Y}}^{t})$.
        \STATE $\boxdot$ \textbf{E-Step: Inference Procedure}
        \STATE Construct the pseudo dataset $(\mathcal{X}^{t+1}, \mathcal{Y}^{t+1})$.
        \STATE Train $p^{AR}(y|x;\phi^{t+1})$ on $(\mathcal{X}^{t+1}, \mathcal{Y}^{t+1})$.
   \ENDWHILE
   \STATE {\bfseries Output:} A NAR model $p^{NAR}(y|x;\theta^{t})$.
\end{algorithmic}
\end{algorithm}

\subsection{The Optimal De-duplicated Decoding Module}

Word duplication is a well-known problem in NAR models caused by the multi-modality issue. To improve the performance of NAR models, some previous work~\citep{lee2018deterministic,li2019hint} remove any duplication in the model prediction by collapsing multiple consecutive occurrences of a token. Such an empirical approach is not technically sound. After collapsing, the length of the target sequence is changed. This will cause a discrepancy between the predicted target length and the actual sequence length and thus make the final output sub-optimal.

We aim to solve the word duplication problem in NAR models, while preserving the original sequence length. Similar to \citet{sun2019fast}, 
we use the Conditional Random Fields (CRF) model ~\citep{lafferty2001conditional} for the decoding of NAR models. The CRF model is manually constructed as follows. It treats the tokens to be decoded as the predicted labels. The unitary scores of the labels in 
each position are set to be NAR models' output distribution and
the transition matrix is set to $-\infty \cdot \mathbf{I}$, where $\mathbf{I}$ is an identity matrix.

Our model is able to find the optimal decoding when considering only the top-3 candidates w.r.t. the unitary scores in each position:
\begin{prop} \label{prop:3}
In a CRF with a transition matrix of $-\infty \cdot \mathbf{I}$, only top-3 likely labels for each position are possible to appear in the optimal (most likely) label sequence.
\end{prop}
We can thus crop the transition matrix accordingly by only keeping a $3\times3$ transition sub-matrix between each pair of adjacent positions.
The forward-backward algorithm \citep{lafferty2001conditional} is then applied on the top-3 likely labels and the $3\times3$ transition sub-matrices to find the optimal decoding with the linear time complexity of $O(|y|)$.

The proposed decoding module is a lightweight plug-and-play module that can be used for any NAR models. Since this principled decoding method is guaranteed to find the optimal prediction that has no word duplication, we refer it as optimal de-duplicated (ODD) decoding method\footnote{Although this method does not allow any word duplication in the output sequence, it is still able to produce any sequence. To solve the problem that multiple consecutive occurrences of a token cannot be captured by our decoding method, we can introduce a special ``$\langle$concat$\rangle$'' symbol. For example, ``very very good'' can be represented by ``very $\langle$concat$\rangle$ very good''.}.

\section{Experiments} \label{sec:experiments}

\subsection{Experimental Settings}

We use several benchmark tasks to evaluate the effectiveness of the proposed method, including IWSLT14\footnote{\url{https://wit3.fbk.eu/}} German-to-English translation (IWSLT14 De-En) and WMT14\footnote{\url{http://statmt.org/wmt14/translation-task.html}}   English-to-German/German-to-English translation (WMT14 En-De/De-En). For the WMT14 dataset, we use Newstest2014 as test data and Newstest2013 as validation data. For IWSLT14/WMT14 datasets, we split words into BPE tokens~\citep{sennrich2015neural}, forming a 10k/32k vocabulary shared by source and target languages.

We use the Transformer~\citep{vaswani2017attention} model as the AR teacher, and the vanilla Non-Autoregressive Transformer (NAT)~\citep{gu2017non} model with sequence-level knowledge distillation~\citep{kim2016sequence} as the NAR baseline.
For both AR and NAR models, we use the original \texttt{base} setting for the WMT14 dataset, and a \texttt{small} setting for the IWSLT14 dataset.
To investigate the influence of the model size on our method, we also evaluate \texttt{large}/\texttt{base} NAT models on WMT14/IWSLT14 datasets as a larger model setting.
These larger NAT models are not used in the EM iterations. They are merely trained with the final AR teacher from the EM iterations of the original model (\texttt{base}/\texttt{small} for WMT14/IWSLT14).
The detailed settings of the model architectures can be founded in the appendix.

We use Adam optimizer~\citep{kingma2014adam} and employ a label smoothing~\citep{szegedy2016rethinking} of 0.1 in all experiments. The \texttt{base} and \texttt{large} models are trained for 125k steps on 8 TPUv3 nodes in each iteration, while the \texttt{small} models are trained for 20k steps. We use a beam size of 20/5 for the AR model in the M/E-step of our EM training algorithm. The pseudo bounds $\{\hat{b}_i\}$ is set by early stopping with the accuracy on the validation set.

\subsection{Inference}

During decoding, the target length $l = |y|$ is predicted by an additional classifier conditional on the source sentence: $l = \argmax_{T'} p(T'|x)$. We can also try different target lengths ranging from $(l - b)$ to $(l + b)$ and obtain multiple translations with different lengths, where $b$ is the half-width, and then use the AR model $p^{AR}(y|x; \phi^1)$ as the rescorer to select the best translation. Such a decoding and rescoring process can be conducted in parallel and is referred as parallel length decoding. To make a fair comparison with previous work, we set $b$ to 4 and use 9 candidate translations for each sentence.
For each dataset, we evaluate the model performance with the BLEU~\citep{papineni2002bleu} score.
\footnote{We follow common practice in previous works to make a fair comparison. Specifically, we use tokenized case-sensitive BLEU for WMT datasets and case-insensitive BLEU for IWSLT datasets.}
We evaluate the average per-sentence decoding latency\footnote{\dag\ in Tab.~\ref{tab2} and Tab.~\ref{tab3} indicates that the latency and the speedup may be affected by hardware settings and are thus not fair for direct comparison.}
on WMT14 En-De test sets with batch size 1 on a single NVIDIA GeForce RTX 2080 Ti GPU by averaging 5 runs.

\setlength{\tabcolsep}{2pt}
\begin{table}[t]
\centering
\caption{Performance of BLEU score on WMT14 En-De/De-En and IWSLT14 De-En tasks for single-pass NAR models. $"/"$ denotes that the results are not reported in the original paper.
Transformer~\citep{vaswani2017attention} results are based on our own reproduction. }
\vspace{5pt}
\resizebox{\columnwidth}{!}{
\small
\begin{tabular}{l l c c c c c}
\toprule
\multicolumn{2}{l}{\multirow{2}{*}{Models}} & \multicolumn{2}{c}{WMT14} & IWSLT14 &  \multirow{2}{*}{Latency} & \multirow{2}{*}{Speedup}\\
\multicolumn{2}{l}{} &En-De & De-En & De-En & &  \\
\hline
\multicolumn{7}{c}{Autoregressive teacher model}\\
\multicolumn{2}{l}{Transformer} & \multirow{2}{*}{27.84} & \multirow{2}{*}{32.14} & \multirow{2}{*}{34.69} & \multirow{2}{*}{393\emph{ms}} & \multirow{2}{*}{1.00$\times$} \\
\multicolumn{2}{l}{\quad w/ beam size 5} & \\
\hline
\multicolumn{7}{c}{Non-autoregressive models}\\
\multicolumn{2}{l}{NAT-FT} & 17.69 & 21.47 &   /   & 39\emph{ms}$^\dag$  & 15.6$\times^\dag$ \\
\multicolumn{2}{l}{LT} & 19.80 &   /   &   /   & 105\emph{ms}$^\dag$ & / \\
\multicolumn{2}{l}{ENAT} & 20.65 & 23.02 & 24.13  & 24\emph{ms}$^\dag$  & 25.3$\times^\dag$ \\
\multicolumn{2}{l}{NAT-BAN} & 21.47 & / & / & / & / \\
\multicolumn{2}{l}{NAT-REG} & 20.65 & 24.77  & 23.89  & 22\emph{ms}$^\dag$ & 27.6$\times^\dag$ \\
\multicolumn{2}{l}{NAT-HINT} & 21.11 & 25.24 & 25.55 & 26\emph{ms}$^\dag$ & 30.2$\times^\dag$\\
\multicolumn{2}{l}{NAT-CTC} & 17.68  & 19.80  &   /   &       350\emph{ms}$^\dag$      & 3.42$\times^\dag$ \\
\multicolumn{2}{l}{FlowSeq-base} & 21.45 & 26.16 & 27.55 & / & /\\
\multicolumn{2}{l}{ReorderNAT} & 22.79 & 27.28 & / & / & 16.1$\times^\dag$\\
\multicolumn{2}{l}{NAT-CRF} & 23.44 & 27.22 & 27.44 & 37\emph{ms}$^\dag$ & 10.4$\times^\dag$\\
\hline
\multicolumn{7}{c}{Ours}\\
\multicolumn{2}{l}{NAT baseline} & 19.55 & 23.44 & 22.35 & 22\emph{ms}& 17.9$\times$\\
\multicolumn{2}{l}{\quad + EM training} & 23.27 & 26.73 & 29.38 & 22\emph{ms}& 17.9$\times$ \\
\multicolumn{2}{l}{\quad + ODD decoding} & \textbf{24.54} & \textbf{27.93} & \textbf{30.69} & 24\emph{ms}& 16.4$\times$\\
\bottomrule
\end{tabular}
}
\label{tab2}
\end{table}
\setlength{\tabcolsep}{6pt}

\setlength{\tabcolsep}{2pt}
\begin{table}[t]
\centering
\caption{Performance of BLEU score on WMT14 En-De/De-En and IWSLT14 De-En tasks for NAR models with rescoring or iterative refinement. $"/"$ denotes that the results are not reported in the original paper.
Transformer~\citep{vaswani2017attention} results are based on our own reproduction. }
\vspace{5pt}
\resizebox{\columnwidth}{!}{
\small
\begin{tabular}{l l c c c c c}
\toprule
\multicolumn{2}{l}{\multirow{2}{*}{Models}} &  \multicolumn{2}{c}{WMT14} & IWSLT14 &  \multirow{2}{*}{Latency} & \multirow{2}{*}{Speedup}\\
\multicolumn{2}{l}{} &En-De & De-En & De-En & &  \\
\hline

\multicolumn{7}{c}{Autoregressive teacher model}\\
\multicolumn{2}{l}{Transformer} & \multirow{2}{*}{27.84} & \multirow{2}{*}{32.14} & \multirow{2}{*}{34.69} & \multirow{2}{*}{393\emph{ms}} & \multirow{2}{*}{1.00$\times$} \\
\multicolumn{2}{l}{\quad w/ beam size 5} & \\
\hline
\multicolumn{7}{c}{Non-autoregressive models}\\
\multicolumn{2}{l}{NAT-FT (rescore 10)} & 18.66  & 22.41  &   /   & 79\emph{ms}$^\dag$  & 7.68$\times^\dag$ \\
\multicolumn{2}{l}{LT (rescore 10)} & 21.00  &   /   &   /   &       /      &       /      \\
\multicolumn{2}{l}{ENAT (rescore 9)} & 24.28 & 26.10 & 27.30  & 49\emph{ms}$^\dag$  & 12.4$\times^\dag$ \\
\multicolumn{2}{l}{NAT-REG (rescore 9)} & 24.61  & 28.90  & 28.04 & 40\emph{ms}$^\dag$ & 15.1$\times^\dag$ \\
\multicolumn{2}{l}{NAT-HINT (rescore 9)} & 25.20 & 29.52 & 28.80 & 44\emph{ms}$^\dag$ & 17.8$\times^\dag$\\
\multirow{2}{*}{FlowSeq} & base (rescore 15) & 23.08 & 28.07 & / & / & 1.04$\times^\dag$ \\
& large (rescore 15) &  25.03 & 30.48 & / & / & 0.96$\times^\dag$\\
\multicolumn{2}{l}{ReorderNAT (rescore 7)} & 24.74 & 29.11 & / & / & 7.40$\times^\dag$\\
\multicolumn{2}{l}{NAT-CRF (rescore 9)} & 26.07 & 29.68 & 29.99 & 63\emph{ms}$^\dag$ & 6.14$\times^\dag$\\
\multicolumn{2}{l}{NAT-IR (10 iterations)} & 21.61  & 25.48  & / & 222\emph{ms}$^\dag$ & 1.88$\times^\dag$ \\
\multirow{2}{*}{CMLM} & (4 iterations) & 25.94 & 29.90 & / & / & 3.05$\times^\dag$ \\
& (10 iterations) & 27.03 & \textbf{30.53} & / & / & 1.30$\times^\dag$ \\
\multicolumn{2}{l}{LevT (7+ iterations)} & \textbf{27.27} & / & / & / & 3.55$\times^\dag$ \\
\hline
\multicolumn{7}{c}{Ours}\\
\multicolumn{2}{l}{NAT baseline (rescore 9)} & 22.24 & 26.21 & 29.64 & 41\emph{ms}& 9.59$\times$\\
\multicolumn{2}{l}{\quad + EM training} & 25.03 & 28.78 & 31.74 & 41\emph{ms}& 9.59$\times$ \\
\multicolumn{2}{l}{\quad + ODD decoding}& 25.75 & 29.29 & 32.66 & 43\emph{ms} & 9.14$\times$ \\
\multicolumn{2}{l}{\quad + larger model}& 26.21 & 29.81 & \textbf{33.25} & 65\emph{ms} & 6.05$\times$\\
\bottomrule
\end{tabular}
}
\label{tab3}
\end{table}
\setlength{\tabcolsep}{6pt}

\subsection{Main Results}
We compare our model with the Transformer~\citep{vaswani2017attention} teacher model and several NAR baselines, including NAT-FT~\citep{gu2017non}, LT~\citep{kaiser2018fast}, ENAT~\citep{guo2018non}, NAT-BAN~\citep{zhou2019understanding}, NAT-REG~\citep{wang2019non}, NAT-HINT~\citep{li2018hint}, NAT-CTC~\citep{libovicky2018end}, Flowseq~\citep{ma2019flowseq}, ReorderNAT~\citep{ran2019guiding}, NAT-CRF~\citep{sun2019fast}, NAT-IR \citep{lee2018deterministic}, CMLM~\citep{ghazvininejad2019mask}, and LevT~\citep{gu2019levenshtein}.

Tab.~\ref{tab2} provides the performance of our method with maximum likely target length $l$, together with other NAR baselines that generate output sequences in a single pass. From the table, we can see that the EM training contributes most to the improvement of the performance. The optimal de-duplicated (ODD) decoding also significantly improves the model performance. Compared with other models, our method significantly outperforms all of them, with nearly no additional overhead compared with the vanilla NAT.

Tab.~\ref{tab3} illustrates the performance of our method equipped with rescoring and other baselines equipped with rescoring or iterative refinement. Since our method has nearly no additional overhead compared with the vanilla NAT, to make a fair comparison with previous work~\citep{kaiser2018fast,lee2018deterministic,ma2019flowseq,sun2019fast,gu2019levenshtein}, we also show the results of our method with a larger model setting. From the table, we can still find that the EM training significantly improves the performance of the vanilla NAT model, but the effect of the OOD decoding is not as significant as in the single-pass setting. This shows that the rescoring process can mitigate the word duplication problem to some extent. Surprisingly, we also find that using the larger model also does not bring much gain. A potential explanation for this is that since our EM algorithm significantly simplifies the training dataset and the NAT model can be over-parameterized, there is no much gain in further increasing the model size. Compared with other baselines, our method significantly outperforms these rescored single-pass NAR methods and achieves competitive performance to iterative-refinement models with a much better speedup. Note that these iterative-refinement models~\citep{lee2018deterministic,ghazvininejad2019mask,gu2019levenshtein} still rely on sequence-level knowledge distillation in training, which indicates that it is still a hard problem for these approaches to capture multi-modality in the real data. Our EM algorithm may further improve their performance. We leave combining the two techniques for future work.

\subsection{Analysis of the Convergence}

We analyze the convergence of the EM algorithm. We show the dynamics of the performance of the NAR model (test BLEU), the performance of the AR model (test BLEU), and the Normalized Corpus-level Multi-modality ($\mathrm{NCM}$, defined by Eq.~\ref{eq:4}) on the WMT14 En-De dataset. The results are shown in Tab.~\ref{tab:converge}.

We describe the detailed optimization process here to clarify how our EM algorithm works with early stopping in this example. In the first 5 iterations, as we have no idea how to set $\{\hat{b}^i\}$ precisely, we simply set them to zeros. But after the $5^{th}$ iteration, we find an accuracy drop in the validation set. So we will re-use the quality metrics at the $4^{th}$ iteration to set $\{\hat{b}^i\}$ and continue the EM algorithm until convergence.

We can see that our EM algorithm takes only a few iterations to convergence, which is very efficient. With the EM algorithm continues, the $\mathrm{NCM}$ metric, which can be regarded as the optimization objective, decreases monotonically. The performance of the NAR model and the performance of the AR model also converge after 5 iterations.

\begin{table}[h]
    \centering
    \small
    \caption{Analysis of the convergence for the EM algorithm on WMT14 En-De test BLEU. $\mathrm{NCM}$ represents the Normalized Corpus-level Multi-modality. All the models are evaluated without ODD decoding and rescoring.
    Iteration $t=1,2,3,4,5^*$ are performed without quality constraints.
    Iteration $t=5,6$ are re-started from $t=4$ with quality constraints.}
    \vspace{5px}
    \begin{tabular}{l c c c}
        \hline
          & NAR Model & AR Model & $\mathrm{NCM}$\\
        \hline
        $t=1$ & 19.55 & \textbf{27.84} & 2.88\\
        $t=2$ & 22.27 & 27.50 & 2.33\\
        $t=3$ & 22.85 & 27.13 & 2.24\\
        $t=4$ & \textbf{23.27} & 26.78 & 2.16\\
        $t=5^*$ & 22.86 & 26.11 & \textbf{2.04}\\
        $t=5$ & 23.18 & 26.72 & 2.12\\
        $t=6$ & 23.16 & 26.75 & 2.11\\
        \hline
    \end{tabular}
    \label{tab:converge}
\end{table}

\subsection{Analysis of the Amortized Inference}

In our EM method, we employ amortized inference and parametrize the posterior distribution of the target sequences by an AR model. In this section, we investigate the importance of amortized inference. Specifically, we try to directly train the NAR model on $(\mathcal{X}^{t+1}, \mathcal{Y}^{t+1})$ in the M-step. The results are shown in Tab.~\ref{tab:amortized}. We can see that parameterizing the posterior distribution by a unified AR model always improves the performance of the NAR model.

\begin{table}[h]
    \centering
    \small
    \caption{Analysis of the amortized inference for iteration $t=2,3,4$ on WMT En-De test BLEU. All the models are evaluated without ODD decoding and rescoring. We show the results of the NAR models using different training data in the M-step.}
    \vspace{5px}
    \begin{tabular}{c c c}
        \hline
         & amortized & non-amortized \\
        \hline
        $t=2$ & \textbf{22.27} & 21.78 \\
        $t=3$ & \textbf{22.85} & 22.44 \\
        $t=4$ & \textbf{23.27} & 22.98\\
        \hline
    \end{tabular}
    \label{tab:amortized}
\end{table}

\subsection{Analysis of the Optimal De-duplicated Decoding}

Finally, we analyze the effect of the proposed optimal de-duplicated (ODD) decoding approach. We compare it with another plug-and-play de-duplicated decoding approach, that is, ``removing any repetition by collapsing multiple consecutive occurrences of a token''~\citep{lee2018deterministic}, which we refer to as post-de-duplication. The results are shown in Tab.~\ref{tab:dedup}. We can see that the proposed ODD decoding method consistently outperforms this empirical method. This shows that our proposed method can overcome the sub-optimal problem of the post-de-duplication method.

\setlength{\tabcolsep}{4pt}
\begin{table}[h]
    \centering
    \small
    \caption{Analysis of the ODD decoding on WMT En-De test BLEU. All the models are trained with our EM algorithm.}
    \vspace{5px}
    \begin{tabular}{l c c c}
        \hline
         & WMT En-De & WMT De-En & IWSLT\\
        \hline
        post-de-duplication & 23.67 & 26.93 & 24.96 \\
        \quad+ rescoring 9& 25.56 & 28.92 & 32.03 \\
        ODD decoding & \textbf{24.54} & \textbf{27.93} & \textbf{30.69}\\
        \quad+ rescoring 9& \textbf{25.75} & \textbf{29.29} & \textbf{32.66}\\
        \hline
    \end{tabular}
    \label{tab:dedup}
\end{table}
\setlength{\tabcolsep}{6pt}

\section{Conclusion}
This paper proposes a novel
EM approach to non-autoregressive conditional sequence generation, which effectively addresses the multi-modality issue in NAR training by iterative optimizing both the teacher AR model and the student NAR model.
We also developed a principled plug-and-play decoding method for efficiently removing word duplication in the model's output.
Experimental results on three tasks prove the effectiveness of our approach.
For future work, we plan to examine the effectiveness of our method in a broader range of applications, such as text summarization.

\section*{Acknowledgements}

We thank the reviewers for their helpful comments.
This work is supported in part by the National Science Foundation (NSF) under grant IIS-1546329.

\bibliography{example_paper}
\bibliographystyle{icml2020}

\onecolumn

\appendix

\newcommand{\vect}[1]{\boldsymbol{#1}}

\section{Non-autoregressive Transformers are Universal Approximators of Sequence-to-Sequence Functions}

\subsection{Problem Definition}

A non-autoregressive Transformer \citep{vaswani2017attention} a Transformer encoder and a non-autoregressive Transformer decoder. More concretely, both encoder and decoder consist of two types of layers: the multi-head attention layer $\mathrm{Attn}$ and the token-wise feed-forward layer $\mathrm{FF}$, with both layers having a skip connection\footnote{The bias $\vect{b}$ is omitted for all matrix multiplication operations for brevity.} \citep{he2016deep}.
The encoder block in the non-autoregressive Transformer ${\rm t}_{\rm enc}$ maps an input $\vect{X} \in \mathbb{R}^{d\times n}$ consisting of $d$-dimensional embeddings of $n$ tokens  to an output ${\rm t}_{\rm enc}(\vect{X}) \in \mathbb{R}^{d\times m}$. It consists of a self-attention layer and a feed-forward layer. The decoder block in the non-autoregressive Transformer ${\rm t}_{\rm dec}$ maps an input $\vect{Y} \in \mathbb{R}^{d\times m}$ consisting of $d$-dimensional embeddings of $m$ tokens and a context $\vect{X} \in \mathbb{R}^{d\times n}$ consisting of $d$-dimensional embeddings of $n$ tokens to an output ${\rm t}_{\rm dec}(\vect{X}, \vect{Y}) \in \mathbb{R}^{d\times m}$. It consists of a self-attention layer, a encoder-decoder attention layer, and a feed-forward layer:
\begin{align}
    \mathrm{Attn}(\vect{X}, \vect{Y}) &= \vect{Y} + \sum_{i=1}^h \vect{W}^i_O\vect{W}^i_V \vect{X} \cdot \sigma\left[(\vect{W}^i_Q \vect{X})^T (\vect{W}^i_K\vect{Y})\right], \\
    \mathrm{FF}(\vect{Y}) &= \vect{Y} + \vect{W}_2 \cdot \mathrm{ReLU}(\vect{W}_1 \cdot \vect{Y}),\\
    {\rm t}_{\rm enc}(\vect{X}) &= \mathrm{FF}(\mathrm{Attn}_{\rm enc-self}(\vect{X}, \vect{X})),\\
    {\rm t}_{\rm dec}(\vect{X}, \vect{Y}) &= \mathrm{FF}(\mathrm{Attn}_{\rm enc-dec}(\vect{X}, \mathrm{Attn}_{\rm dec-self}(\vect{Y}, \vect{Y}))),
\end{align}
where $\vect{W}^i_O \in \mathbb{R}^{d\times k}$, $\vect{W}^i_V, \vect{W}^i_K, \vect{W}^i_Q  \in \mathbb{R}^{k\times d}$, $\vect{W}_2 \in \mathbb{R}^{d\times r}$, and $\vect{W}_1 \in \mathbb{R}^{r\times d}$ are learnable parameters. $\sigma$ is the softmax function. Following \citet{yun2020are}, we also do not use layer normalization \citep{ba2016layer} in the setup of our analysis.

The family of the Transformer encoders is $\mathbb{R}^{d \times n} \rightarrow \mathbb{R}^{d \times n}$ functions and can be defined as:
\begin{equation} \label{eq:tfe}
    \mathcal{T}_{\rm enc}^{h, k, r}\coloneqq \left\{
    h: \mathbb{R}^{d \times n} \rightarrow \mathbb{R}^{d \times n} \middle\vert\
    \begin{split}
    \vect{X}^0 &= \vect{X}\\
    \vect{X}^i &= {\rm t}_{\rm enc}^{h,k,r}(\vect{X}^{i-1})\\
    h(\vect{X}) &= \vect{X}^M
    \end{split}
    \right\},
\end{equation}
where ${\rm t}_{\rm enc}^{h,k,r}$ denotes a Transformer encoder block defined by an attention layer with $h$ heads of size $k$ each, and a feed-forward layer with $r$ hidden nodes. $M$ is the number of stacked blocks.

Similarly, the family of the non-autoregressive Transformer decoders is $\mathbb{R}^{d \times (n + m)} \rightarrow \mathbb{R}^{d \times m}$ functions and can be defined as:
\begin{equation} \label{eq:tfd}
    \mathcal{T}_{\rm dec}^{h, k, r}\coloneqq \left\{
    h: \mathbb{R}^{d \times (n+m)} \rightarrow \mathbb{R}^{d \times m} \middle\vert\
    \begin{split}
    \vect{Y}^0 &= \vect{Y}\\
    \vect{Y}^i &= {\rm t}_{\rm dec}^{h,k,r}(\vect{X}, \vect{Y}^{i-1})\\
    h(\vect{X}, \vect{Y}) &= \vect{Y}^N
    \end{split}
    \right\},
\end{equation}
where ${\rm t}^{h,k,r}_{\rm dec}$ denotes a Transformer decoder block defined by attention layers with $h$ heads of size $k$ each and a feed-forward layer with $r$ hidden nodes. $N$ is the number of stacked blocks.

Finally, the family of non-autoregressive Transformers is $\mathbb{R}^{d \times n} \rightarrow \mathbb{R}^{d \times m}$ functions and can be defined as:
\begin{equation} \label{eq:tf}
    \mathcal{T}^{h, k, r}\coloneqq \left\{ g(\vect{X}) = h_2(h_1(\vect{X} + \vect{E}_1), \vect{E}_2) \middle\vert\ h_1 \in \mathcal{T}_{\rm enc}^{h, k, r} \text{ and } h_2 \in \mathcal{T}_{\rm dec}^{h, k, r}
    \right\},
\end{equation}
where $\vect{E}_1 \in \mathbb{R}^{d\times n}$ and $\vect{E}_2 \in \mathbb{R}^{d\times m}$ are the trainable positional embeddings.

\subsection{Transformer Encoders are Universal Approximators of Sequence-to-Sequence Functions \citep{yun2020are}}

Recently, \citet{yun2020are} show that the Transformer encoders equipped with positional embeddings are universal approximators of all continuous $\mathbb{R}^{d\times n} \rightarrow \mathbb{R}^{d\times n}$ functions that map a compact domain in $\mathbb{R}^{d\times n}$ to $\mathbb{R}^{d\times n}$.

We first describe the results in \citet{yun2020are}. Let us start by defining the target function class $\mathcal{F}_{\rm enc}$, which consists of all continuous sequence-to-sequence functions with compact support that map a compact domain in $\mathbb{R}^{d\times n}$ to $\mathbb{R}^{d\times n}$. Here continuity is defined with respect to any entry-wise $\ell^p$ norm, $1 \le p < \infty$. Given two functions $f_1, f_2: \mathbb{R}^{d\times n} \rightarrow \mathbb{R}^{d\times n}$, for $1 \le p < \infty$, we define a distance between them as
\begin{equation} \label{eq:dis}
    d_p(f_1, f_2) \coloneqq \left(\int \|f_1(\vect{X}) - f_2(\vect{X})\|^p_p d\vect{X}\right)^{1/p}.
\end{equation}
The Transformer encoders with positional embeddings is defined as:
\begin{equation}
    \mathcal{T}^{h,k,r}_{\rm P-enc} \coloneqq \left\{h_{\rm P}{\vect{X}} = h(\vect{X} + \vect{E}) | h \in \mathcal{T}_{\rm enc}^{h, k, r} \text{ and } \vect{E} \in \mathbb{R}^{d\times n} \right\},
\end{equation}
where $\vect{E}$ is learnable positional embeddings. The following result shows that a Transformer encoder with positional embeddings with a constant number of heads $h$, head size $k$, and hidden layer of size $r$ can approximate any function in $\mathcal{F}_{\rm enc}$:
\begin{theorem} \label{the:enc}
Let $1 \le p < \infty$ and $\epsilon > 0$, then for any given $f \in \mathcal{F}_{\rm enc}$, there exists a Transformer encoder $h \in \mathcal{T}^{2,1,4}_{\rm P-enc}$ such that we have $d_p(f,h) \le \epsilon$.
\end{theorem}
We provide the sketch of the proof in \citet{yun2020are} here. Without loss of generality, we can assume that the compact support of $f$ is contained in $[0, 1]^{d\times n}$. The proof of Theorem~\ref{the:enc} can be achieved in the following three steps:
\paragraph{Step 1: Approximate $\mathcal{F}_{\rm enc}$ with piece-wise constant functions.} We first use (a variant of) the classical result that any continuous function can be approximated up to arbitrary accuracy by piece-wise constant functions. For $\delta > 0$, we define the following class of piece-wise constant functions:
\begin{equation}
\overline{\mathcal{F}}_{\rm enc}(\delta) \coloneqq
    \left\{ f: \vect{X} \mapsto \sum\nolimits_{\vect{L} \in \mathbb{G}_{\delta}} 
    \vect{A}_{\vect{L}} 
    \mathbbm{1}\{\vect{X} \in \mathbb{G}_{\vect{L}}\}
    \middle\vert\
    \vect{A}_{\vect{L}} \in \mathbb{R}^{d\times n}
    \right\},   
\end{equation}
where $\mathbb{G}_{\delta} \coloneqq \{0, \delta, \dots, 1-\delta \}^{d \times n}$ and, for a grid point $\vect{L} \in \mathbb{G}_{\delta}$, $\mathbb{S}_{\vect{L}} \coloneqq \prod_{j=1}^d \prod_{k=1}^n [L_{j,k}, L_{j,k}+\delta) \subset [0,1]^{d \times n}$ denotes the associated cube of width $\delta$. Let $\overline{f} \in \overline{\mathcal F}_{\rm enc}(\delta)$ be such that $d_{p}(f, \overline{f}) \leq \epsilon/3$. 

\paragraph{Step 2: Approximate $\overline {\mathcal F}_{\rm enc} (\delta)$ with {\em modified} Transformer encoders.}
We then consider a slightly modified architecture for Transformer networks, where the softmax operator $\sigma[\cdot]$ and $\mathrm{ReLU}(\cdot)$ are replaced by the hardmax operator $\sigma_{\mathrm H}[\cdot]$ and an activation function $\phi \in \Phi$, respectively. 
Here, the set of allowed activations $\Phi$ consists of all piece-wise linear functions with at most three pieces, where at least one piece is constant.
Let $\overline{\mathcal T}^{h, k, r}_{\mathrm enc}$ denote the function class corresponding to the sequence-to-sequence functions defined by the modified Transformer encoders. The following result establishes that the modified Transformer encoders in $\overline{\mathcal T}^{2, 1, 1}_{\mathrm enc}$ can closely approximate functions in $\overline {\mathcal F}_{\rm enc}(\delta)$.

\begin{prop}
\label{prop:part2}
For each $\overline f \in \overline {\mathcal F}_{\rm enc} (\delta)$ and $1\leq p < \infty$, $\exists$ $\overline g \in \overline {\mathcal T}^{2,1,1}_{\rm enc}$ such that $d_p(\overline f, \overline g) = O(\delta^{d/p})$.
\end{prop}

\paragraph{Step 3: Approximate {\em modified} Transformer encoders with (original) Transformer encoders.}
Finally, we show that $\overline{g} \in \overline {\mathcal T}^{2,1,1}$ can be approximated by $\mathcal T^{2, 1, 4}$. 
Let $g \in {\mathcal T}^{2,1,4}$ be such that $d_{p}(\overline{g}, g)\leq \epsilon/3$.

Theorem~\ref{the:enc} now follows from these three steps, because we have
\begin{equation}
    d_p(f, g) \le d_p(f, \overline{f}) + d_p(\overline{f}, \overline{g}) + d_p(\overline{g}, g) \le 2\epsilon/3 + O(\delta^{d/p}).
\end{equation}
Choosing $\delta$ small enough ensures that $d_p(f, g) \leq \epsilon$. 

\subsection{Proof Sketch of Proposition~\ref{prop:part2} \citep{yun2020are}}

Especially, the proof of Proposition~\ref{prop:part2} is decomposed into three steps:
\paragraph{Sub-step 1: Quantization by feed-forward layers}
Given an input $\mX \in \mathbb{R}^{d \times n}$, a series of feed-forward layers in the modified Transformer encoder can quantize $\mX$ to an element $\mL$ on the extended grid 
$\mathbb{G}^+_{\delta} \coloneqq \{ -\delta^{-nd}, 0, \delta, \dots, 1-\delta \}^{d \times n}$.
\paragraph{Sub-step 2: Contextual mapping by self-attention layers}
Next, a series of self-attention layers in the modified Transformer encoder can take the input $\mL$ and implement a {\em contextual mapping} $q :\mathbb{G}_{\delta} \rightarrow \mathbb{R}^n$ such that, for $\mL$ and $\mL'$ that are not permutation of each other, all the elements in $q(\mL)$ and $q(\mL')$ are distinct.
\paragraph{Sub-step 3: Function value mapping by feed-forward layers}
Finally, a series of feed-forward layers in the modified Transformer encoder can map elements of the contextual embedding $q(\mL)$ to the desired output value of $\overline{f} \in \overline{\mathcal F}_{\rm enc}$ at the input $\mX$.

\subsection{Non-autoregressive Transformers are Universal Approximators of Sequence-to-Sequence Functions}

In this paper, we take a further step and show that the non-autoregressive Transformers are universal approximators of all continuous $\mathbb{R}^{d\times n} \rightarrow \mathbb{R}^{d\times m}$ functions that map a compact domain in $\mathbb{R}^{d\times n}$ to $\mathbb{R}^{d\times m}$, where $n$ and $m$ can be different.

We start with describing the formal form of Theorem 4.1 in the main text. In the non-autoregressive conditional sequence generation problem, the target function class $\mathcal{F}_{s2s}$ becomes the set of all continuous sequence-to-sequence functions with compact support that map a compact domain in $\mathbb{R}^{d\times n}$ to $\mathbb{R}^{d\times m}$, where $n$ and $m$ can be different. Given two functions $f_1, f_2: \mathbb{R}^{d\times n} \rightarrow \mathbb{R}^{d\times m}$, for $1 \le p < \infty$, similar to Eq.~\ref{eq:dis}, we define a distance between them as
\begin{equation}
    d_p(f_1, f_2) \coloneqq \left(\int \|f_1(\vect{X}) - f_2(\vect{X})\|^p_p d\vect{X}\right)^{1/p}.
\end{equation}
With the definition of non-autoregressive Transformers in Eq.~\ref{eq:tf}, we have the following result:
\begin{theorem} \label{the:nat}
Let $1 \le p < \infty$ and $\epsilon > 0$, then for any given $f \in \mathcal{F}_{\rm s2s}$, there exists a non-autoregressive Transformer $g \in \mathcal{T}^{2,1,4}$ such that we have $d_p(f,g) \le \epsilon$.
\end{theorem}

The proof of Theorem~\ref{the:nat} can be done in a similar way as Theorem~\ref{the:enc}. Especially, the step 1 and step 3 in the proof of Theorem~\ref{the:enc} can be seamlessly used here. We refer the readers to \citet{yun2020are} for the detailed proof of these two steps.

The step 2 in the proof Theorem~\ref{the:nat} is a bit different. Basically, we need to prove the following result:
\begin{prop}
\label{prop:part3}
For each $\overline f \in \overline {\mathcal F}_{\rm s2s} (\delta)$ and $1\leq p < \infty$, $\exists$ $\overline g \in \overline {\mathcal T}^{2,1,1}$ such that $d_p(\overline f, \overline g) = O(\delta^{d/p})$.
\end{prop}
where $\overline {\mathcal F}_{\rm s2s} (\delta)$ and $\overline {\mathcal T}^{2,1,1}$ are defined in a similar way as $\overline {\mathcal F}_{\rm enc} (\delta)$ and $\overline {\mathcal T}^{2,1,1}_{\rm enc}$, respectively.
Similar to the proof of Proposition \ref{prop:part2}, the proof of Proposition \ref{prop:part3} can be decomposed into three steps:
\paragraph{Sub-step 1$^*$: Quantization by feed-forward layers in the encoder}
Given an input $\mX \in \mathbb{R}^{d \times n}$, a series of feed-forward layers in {\em the encoder of the modified non-autoregressive Transformer} can quantize $\mX$ to an element $\mL$ on the extended grid 
$\mathbb{G}^+_{\delta} \coloneqq \{ -\delta^{-nd}, 0, \delta, \dots, 1-\delta \}^{d \times n}$.
\paragraph{Sub-step 2$^*$: Contextual mapping by attention layers in the encoder and the decoder}
Next, a series of attention layers in {\em the encoder and decoder of the modified non-autoregressive Transformer} can take the input $\mL$ and implement a {\em contextual mapping} $q : \mathbb{G}_{\delta} \rightarrow \mathbb{R}^m$ such that, for $\mL$ and $\mL'$ that are not permutation of each other, all the elements in $q(\mL)$ and $q(\mL')$ are distinct.
\paragraph{Sub-step 3$^*$: Function value mapping by feed-forward layers in the decoder}
Finally, a series of feed-forward layers in {\em the decoder of the modified non-autoregressive Transformer} can map elements of the contextual embedding $q(\mL)$ to the desired output value of $\overline{f} \in \overline{\mathcal F}_{\rm s2s}$ at the input $\mX$.

Since Sub-step 1$^*$ and Sub-step 3$^*$ are exactly the same as Sub-step 1 and Sub-step 3 in the proof of Proposition~\ref{prop:part2}, we only provide the proof of Sub-step 2$^*$. We refer the readers to \citet{yun2020are} for the detailed proof of these two sub-steps.

\subsection{Proof of Sub-step 2$^*$ in Proposition~\ref{prop:part3}}

Without loss of generality, we can assume that the compact support of $f$ is contained in $[0, 1]^{d\times n}$. Following \citet{yun2020are}, we choose
\begin{equation*}
    \mE_1 = 
    \begin{bmatrix}
    0 & 1 & 2 & \cdots & n-1\\
    0 & 1 & 2 & \cdots & n-1\\
    \vdots & \vdots & \vdots & & \vdots\\
    0 & 1 & 2 & \cdots & n-1
    \end{bmatrix}.
\end{equation*}
and 
\begin{equation*}
    \mE_2 = 
    \begin{bmatrix}
    0 & 1 & 2 & \cdots & m-1\\
    0 & 1 & 2 & \cdots & m-1\\
    \vdots & \vdots & \vdots & & \vdots\\
    0 & 1 & 2 & \cdots & m-1
    \end{bmatrix}.
\end{equation*}

By Sub-step~1$^*$, we quantize any input $\mX + \mE_1$ to its quantized version with the feed-forward layers in the Transformer encoder. We call this quantized version $\mL$:
\begin{equation*}
\mL \in [0:\delta:1-\delta]^d \times [1:\delta:2-\delta]^d \times \dots \times [n-1:\delta:n-\delta]^d.
\end{equation*}
We do not need to quantize $\mE_2$ in our Sub-step~1$^*$ with the feed-forward layers in the Transformer decoder because $\mE_2$ is already quantized.

As done in Lemma 6 in \citet{yun2020are}, we define $\vu \coloneqq (1, \delta^{-1}, \dots, \delta^{-d+1})$ and $l_j \coloneqq \vu^T \mL_{:,j}$, for all $j \in [n]$. Next, following the construction in Appendix C.2 in \citet{yun2020are}, with $n(1/\delta)^d$ self-attention layers in the Transformer encoder, we can get $\widetilde l_1 < \widetilde l_2 < \cdots < \widetilde l_n$ such that the map from $\mathcal{L}$ to $\widetilde l_n$ is one-to-one. In addition,  $\widetilde l_n$ is bounded by $n\delta^{-(n+1)d}$.

Finally, in a similar way as Appendix B.5.1 in \citet{yun2020are}, we add an extra encoder-decoder attention layer with attention part $n\delta^{-(n+1)d -1} \Psi(\cdot; 0)$. This layer shifts all the layers in the Transformer decoder by $n\delta^{-(n+1)d -1} \widetilde l_n$. We define the output of this layer as $g_c(\vect{L})$. In this way, we ensure that different contexts $\vect{L}$ are mapped to distinct numbers in $\vect{u}^Tg_c(\vect{L})$, thus implementing a contextual mapping.

\section{Proof of Proposition~5.1}

We prove this proposition by contradiction. Assuming that the $k$-th likely label in position $i$ is chosen by the CRF algorithm and $k > 3$, we consider the following two cases:

\paragraph{Case 1: $i = 0$ is the first position or $i=n-1$ is the last position.} Without loss of generality, we can assume $i = 0$. The first and second labels in the current decoding is denoted by $l_0^*$ and $l_1^*$. We also denote the top 2 labels in position $0$ as $l_{0, 1}$ and $l_{0, 2}$.
If $l_{0, 1} = l_1^*$, we can set $l_0^*$ to be $l_{0, 2}$, which construct a label sequence with higher likelihood in the CRF model. Otherwise, we can set $l_0^*$ to be $l_{0, 1}$. In both cases, $k > 3$ is not the optimal solution.

\paragraph{Case 2: $i$ is the neither the first position nor the last position.} We denote the labels on the position before and after $i$ as $l_{i-1}^*$ and $l_{i+1}^*$. We denote the $j$-th likely label on the position $i$ as $l_{i, j}$.
In this case, we will always find such a $j \le 3$ that $l_{i, j} \neq l_{i-1}^*$ and $l_{i, j} \neq l_{i+1}^*$. Therefore, $k > 3$ is still not the optimal solution.
\qed

\section{Illustration of Different Decoding Approaches}

\begin{figure}[t]
    \centering
    \includegraphics[width=0.7\linewidth]{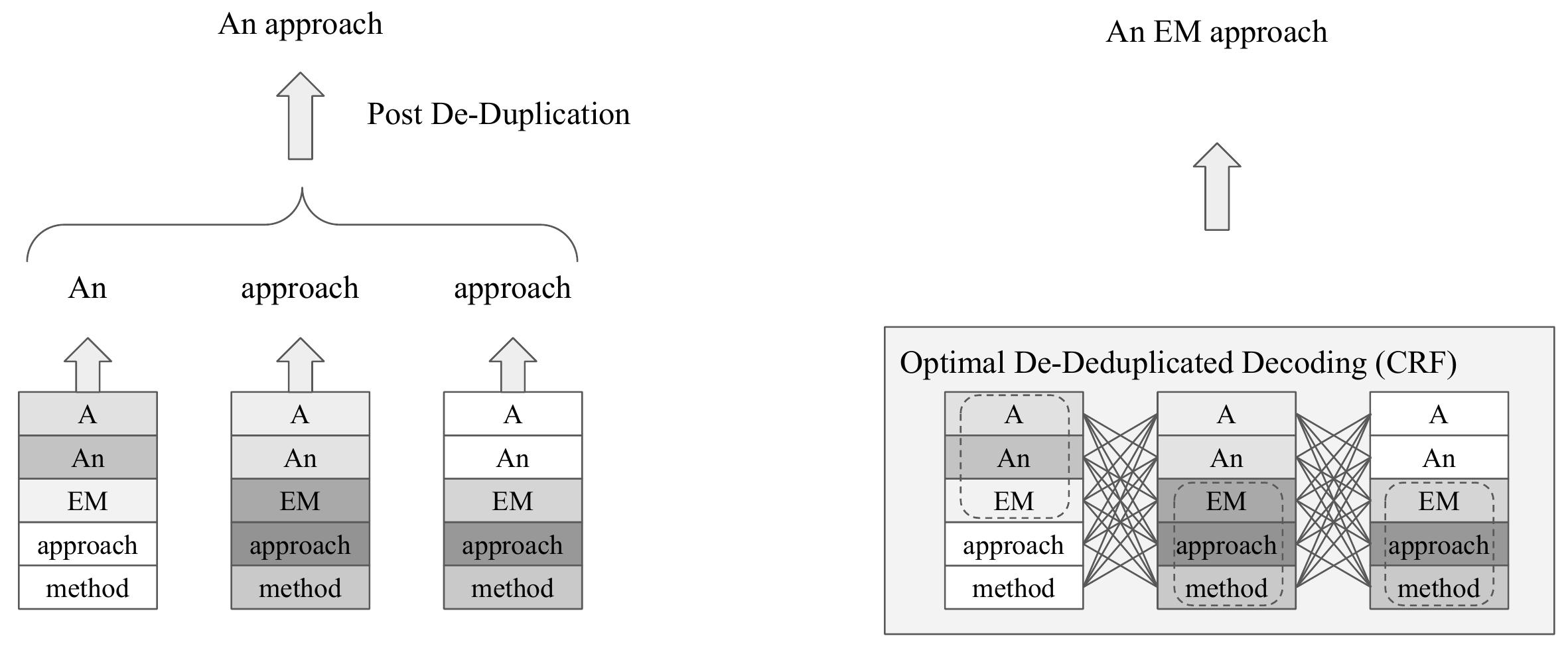}
    \caption{Illustration of different decoding methods. Darker boxes represent higher likelihood. The argmax decoding method produces ``An approach appraoch'' as the result, which contains word duplication. The empirical post de-duplication method can solve the word duplication problem, but after collapsing, the length of the target sequence is changed. This will cause a discrepancy between the predicted target length and the actual sequence length and thus make the final output sub-optimal. The proposed Optimal De-Duplicated (ODD) decoding can produce the optimal prediction in the CRF framework. Note that OOD decoding only needs to consider the top-3 labels for each position in the forward-backward algorithm, which is very efficient.}
    \label{fig:my_label}
\end{figure}

Fig.~\ref{fig:my_label} shows how the proposed optimal de-duplicated decoding method solves the sub-optimal decoding problem of the post de-duplication method.

\section{Model Settings}

\begin{figure}[t]
    \centering
    \includegraphics[width=0.8\linewidth]{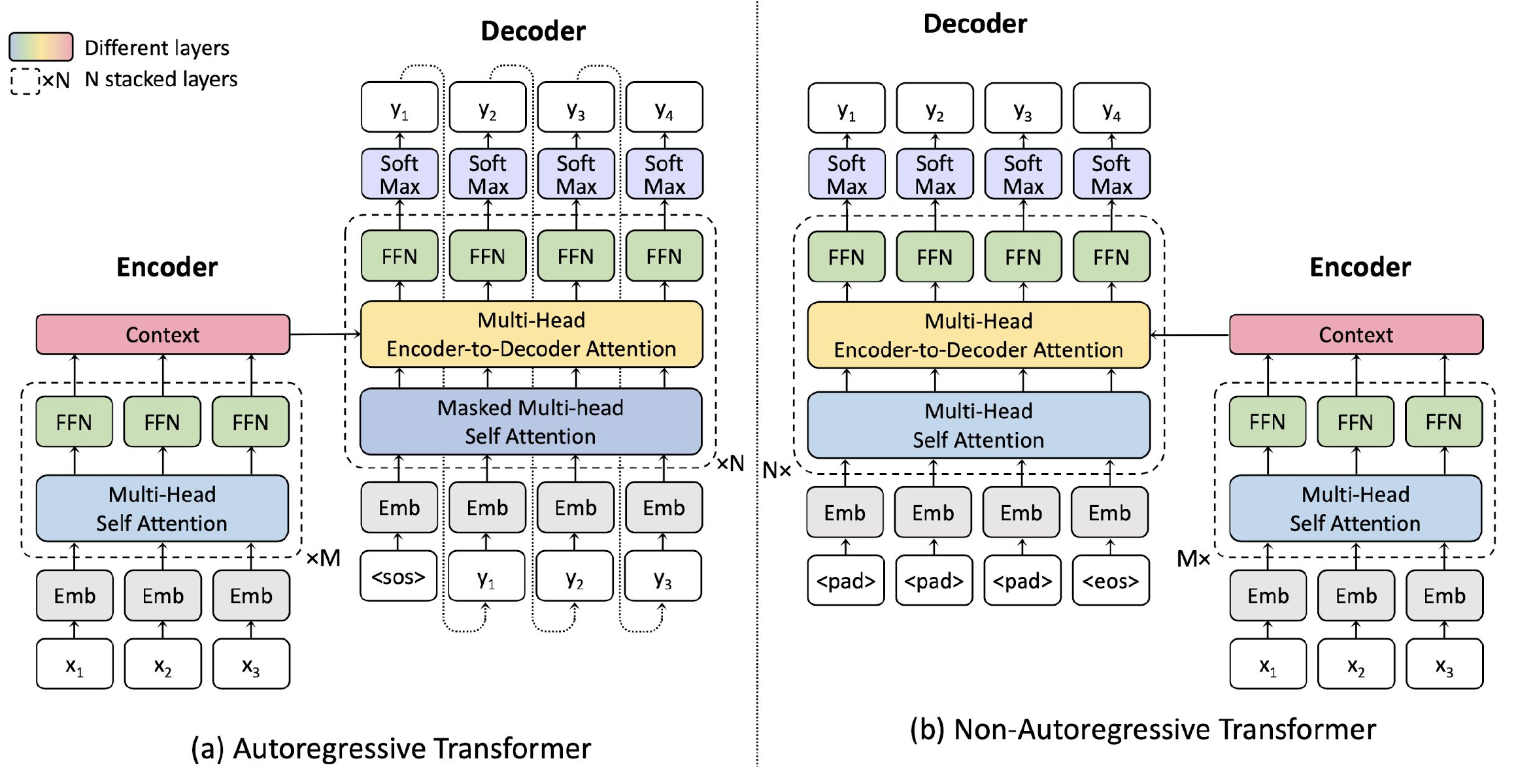}
    \caption{The architectures of Autoregressive Transformer and Non-autoregressive Transformer used in this paper.}
    \label{fig:model}
\end{figure}

The Non-Autoregressive Transformer model (NAT) is developed using the general encoder-decoder framework which is the same as the Autoregressive Transformer (AT) model. Fig.~\ref{fig:model} shows the architectures of NAT and AT. We use a simplified version of the NAT model in \citet{gu2017non}, that is, we do not copy the source embeddings as the input of the Transformer decoder and do not use the positional attention proposed in \citet{gu2017non}. The input to our Transformer decoder is simply the padding symbols.
More details about the description about the architectures can be found in \citet{vaswani2017attention,gu2017non}.

We use four model settings in our experiments, including \texttt{toy}, \texttt{small}, \texttt{base}, and \texttt{large}. The detailed configurations of these four model settings can be found in Tab.~\ref{tab:my_label}.

\begin{table}[b]
    \centering
    \caption{Transformer model settings}
    \begin{tabular}{l c c c c c}
    \hline
         &  encoder-layer & decoder-layer & hidden-size & filter-size & num-heads\\
    \hline
         \texttt{toy} & 3 & 3 & 256 & 1024 & 4\\
         \texttt{small} & 5 & 5 & 256 & 1024 & 4\\
         \texttt{base} & 6 & 6 & 512 & 2048 & 8\\
         \texttt{large} & 6 & 6 & 1024 & 4096 & 16\\
    \hline
    \end{tabular}
    \label{tab:my_label}
\end{table}

\section{Analysis of Training Examples for the NAR Model}\label{appendix:sample}

\newcommand{\gn}[1]{\textcolor{magenta}{\bf\small [#1 --GN]}}
\newcommand{\ct}[1]{\textcolor{blue}{\bf\small [#1 --CZ]}}
\newcommand{\gnc}[2]{\textcolor{magenta}{\sout{#1} #2}}
\definecolor{mypink}{cmyk}{0, 0.7808, 0.4429, 0.1412}
\definecolor{mypurple}{rgb}{0.57, 0.36, 0.51}
\definecolor{myblue}{rgb}{0.2, 0.2, 0.6}

\begin{table*}[ht!]
\centering
\small
\begin{tabular}{p{2.5cm}p{12cm}}
\toprule

\textcolor{myblue}{Source} & \textcolor{myblue}{, um den Korb zu verkleinern ( bis 75 \% ) und in die Ecke zu schieben .} \\
\textcolor{mypink}{Ground Truth} & 
\textcolor{mypink}{to resize the fragment ( by 75 \% ) and move it to the lower right corner .} \\
Iteration 0 &  
to reduce the basket ( up to 75 \% ) and move it to the corner . \\
Iteration 1 &  to reduce the basket ( up to 75 \% ) and push it into the corner . \\
Iteration 2 & to reduce the basket ( up to 75 \% ) and put it into the corner . \\
\midrule
\textcolor{myblue}{Source} & \textcolor{myblue}{In den Interviews betonten viele Männer , dass ihre Erwerbsabweichung ihre Karriere behindere .} \\
\textcolor{mypink}{Ground Truth} & 
\textcolor{mypink}{In the interviews , many men emphasized that their employment deviation has hindered their careers .} \\
Iteration 0 &  
In the interviews , many men emphasized that their divorce in employment hinders their careers . \\
Iteration 1 & In the interviews , many men stressed that their deviation from employment hindered their careers . \\
Iteration 2 & In the interviews , many men stressed that their deviation in employment hinders their careers . \\
\midrule
\textcolor{myblue}{Source} & \textcolor{myblue}{Um einen Pferd gesund und munter zu halten , müssen Sie seine physischen Bedürfnisse beachten .} \\
\textcolor{mypink}{Ground Truth} & 
\textcolor{mypink}{To keep your horse well , healthy and content you must satisfy its physical needs .} \\
Iteration 0 &  
In order to keep a horse healthy and healthy , you must take into account its physical needs . \\
Iteration 1 & In order to keep a horse healthy and cheerful , you must take into account its physical needs . \\
Iteration 2 & In order to keep a horse healthy and cheerful , you must take into account your physical needs . \\
\midrule
\textcolor{myblue}{Source} & \textcolor{myblue}{Der Effekt von \&apos; Eiskältefalle \&apos; wird nun bei erlittenem Schaden abgebrochen .} \\
\textcolor{mypink}{Ground Truth} & 
\textcolor{mypink}{Freezing Trap now breaks on damage .} \\
Iteration 0 &  
Ice Cat Trap effect will now be discarded if damage is dealt . \\
Iteration 1 & The effect of Ice Cage Trap will now be aborted in case of damage suffered . \\
Iteration 2 & The effect of ice cold trap is now aborted in the event of damage suffered . \\
\midrule
\textcolor{myblue}{Source} & \textcolor{myblue}{Wir haben in der Europäischen Union Möglichkeiten , wirksam gegen die Arbeitslosigkeit vorzugehen und zwar so , daß man da , wo es am nötigsten ist , auch etwas davon spürt .} \\
\textcolor{mypink}{Ground Truth} & 
\textcolor{mypink}{We have the opportunity , in the EU , to do something that will have a positive effect on unemployment , characterised by taking action where there is the greatest need .} \\
Iteration 0 &  
We in the European Union have the means to combat unemployment effectively , in such a way that we feel something of it where it is most necessary . \\
Iteration 1 & We in the European Union have opportunities to take effective action against unemployment , in such a way that we can feel something about it where it is most necessary . \\
Iteration 2 & We in the European Union have opportunities to take effective action against unemployment , in such a way that we also feel something of it where it is most necessary . \\
\midrule
\textcolor{myblue}{Source} & \textcolor{myblue}{In der Altstadt sind die Gassen so eng und verwinkelt , dass ein Auto nur mühsam vorankommt .} \\
\textcolor{mypink}{Ground Truth} & 
\textcolor{mypink}{Kaneo Settlement - Start the walk to Kaneo from St. Sophia church .} \\
Iteration 0 &  
In the old town , the streets are so narrow and winding that a car can only progress with difficulty . \\
Iteration 1 & In the old town , the streets are so narrow and winding that a car is progressing hard . \\
Iteration 2 & In the old town , the streets are so narrow and winding that a car is only progressing laboriously . \\
\midrule
\textcolor{myblue}{Source} & \textcolor{myblue}{Sie müssen sich nur einmal die weltweit steigende Anzahl und Häufigkeit von Naturkatastrophen ansehen , um die Folgen der Klimaveränderung zu erkennen .} \\
\textcolor{mypink}{Ground Truth} & 
\textcolor{mypink}{They only need to look at the increasing number and frequency of natural disasters worldwide to see its impact .} \\
Iteration 0 &  
You only have to look at the increasing number and frequency of natural disasters worldwide to see the consequences of climate change . \\
Iteration 1 & They only have to look at the increasing number and frequency of natural disasters around the world to see the consequences of climate change . \\
Iteration 2 & They only have to look at the increasing number and frequency of natural disasters around the world in order to identify the consequences of climate change . \\
\bottomrule
\end{tabular}
\caption{Examples in the training data for the NAR model on the WMT14 De-En task.}
\label{tab:example}
    \vspace{-2mm}
\end{table*}
In Tab. \ref{tab:example}, we present randomly picked examples from the training data for the NAR model on the WMT14 De-En task. We can find that the proposed EM algorithm constantly changes the training examples for the NAR model.

\section{Analysis of Translation Results}

In Tab.~\ref{tab:example2}, we present randomly picked translation outputs from the test set of WMT14 De-En. We have the following observations:
\begin{itemize}
    \item The proposed OOD decoding method preserves the original predicted length of tokens, which avoid the sub-optimal problem of the post de-duplication method.
    \item The proposed EM algorithm can effectively jointly optimize both the AR model and the NAR model. During EM iterations, the multi-modality in the AR models is reduced, while the translation quality of the NAR models is improved.
\end{itemize}

\begin{table*}[ht!]
\centering
\small
\begin{tabular}{p{6cm}p{10cm}}
\toprule

\textcolor{myblue}{Source} & \textcolor{myblue}{Sie ist die Tochter von Peter Tunks , einem ehemaligen Spieler der australischen Rubgy @-@ Liga , der sich an das Außenministerium in Canberra mit der Bitte um Hilfe für seine Tochter gewandt hat .} \\
\textcolor{mypink}{Ground Truth} & 
\textcolor{mypink}{She is the daughter of former Australian league player Peter Tunks , who has appealed to the Department of Foreign Affairs in Canberra to assist his daughter .} \\
AR model - Iter 0 &  
She is the daughter of Peter Tunks , a former player of the Australian Rubgy League , who has addressed to the Ministry of Foreign Affairs in Canberra asking for help for his daughter . \\
AR model - Iter 2 &
She is the daughter of Peter Tunks , a former player of the Australian Rubgy League who addressed the Ministry of Foreign Affairs in Canberra asking for help for his daughter .\\
NAR model - Iter 0& She is the daughter of Peter Tunks , a former player of the Australian Rubgy League , who addressed the Ministry of Foreign Affairs Canberberra asking help help his daughter .\\
NAR model - Iter 0 w/ post de-duplication& She is the daughter of Peter Tunks , a former player of the Australian Rubgy League , who addressed the Ministry of Foreign Affairs Canberra asking help his daughter .\\
NAR model - Iter 2& She is the daughter of Peter Tunks , a former player of the Australian Rubgy League ague who addressed the Ministry of Foreign Affairs in Canberra ra asking for help to his daughter .\\
NAR model - Iter 2 w/ post de-duplication& She is the daughter of Peter Tunks , a former player of the Australian Rubgy League ague who addressed the Ministry of Foreign Affairs in Canberra asking for help to his daughter .\\
NAR model - Iter 2 w/ ODD decoding & She is the daughter of Peter Tunks , a former player of the Australian Rubgy League ague who addressed the Ministry of Foreign Affairs in Canberra for asking for help to his daughter .\\
\midrule

\textcolor{myblue}{Source} & \textcolor{myblue}{In australischen Berichten war zu lesen , dass sie in der Zwischenzeit im Ferienort Krabi in Südthailand Urlaub macht .} \\
\textcolor{mypink}{Ground Truth} & 
\textcolor{mypink}{Reports in Australia said that in the meantime , she was holidaying at the resort area of Krabi in Southern Thailand .} \\
AR model - Iter 0 &  
In Australian reports it was read that in the meantime it is making a holiday in the holiday resort of Krabi in South thailand . \\
AR model - Iter 2 &  Australian reports read that , in the meantime , it is a holiday in the resort of Krabi in southern Thailand .\\
NAR model - Iter 0& Australian reports have that , in the meantime , they is a holiday holiday southern southern in the Krabi of Krabi .\\
NAR model - Iter 0 w/ post de-duplication& Australian reports have that , in the meantime , they is a holiday southern in the Krabi of Krabi .\\
NAR model - Iter 2& Australian reports read that , in the meantime , it is a holiday holiday the resort of Kraresort in southern Thailand .\\
NAR model - Iter 2 w/ post de-duplication & Australian reports read that , in the meantime , it is a holiday the resort of Kraresort in southern Thailand .\\
NAR model - Iter 2 w/ ODD decoding & Australian reports read that , in the meantime , it is a holiday in the resort of Kraresort in southern Thailand .\\
\bottomrule
\end{tabular}
\caption{Examples of translation outputs on the WMT14 De-En task. We do not apply rescoring to the NAR model's outputs.}
\label{tab:example2}
    \vspace{-2mm}
\end{table*}

\end{document}